\documentclass[runningheads]{llncs}
\usepackage{cite}
\usepackage{amsmath,amssymb,amsfonts}
\usepackage{algorithm}
\usepackage{algorithmic}
\usepackage{graphicx}
\usepackage{textcomp}
\usepackage{xcolor}
\usepackage{booktabs}
\usepackage{multirow}
\usepackage{tabularx}
\usepackage{makecell}
\begin{document}
\title{The Power of Adaptation: Boosting In-Context Learning through Adaptive Prompting}
\author{Shuzhang Cai\inst{1} \and
        Twumasi Mensah-Boateng\inst{2} \and
        Xander Kuksov\inst{2} \and
        Jing Yuan\inst{2} \and
        Shaojie Tang\inst{3}}
\authorrunning{S. Cai et al.}
\institute{University of Texas at Dallas, USA \and
           University of North Texas, USA \and
           University at Buffalo, USA}

\maketitle              
\begin{abstract}
Large Language Models (LLMs) have demonstrated exceptional abilities across a broad range of language-related tasks, including generating solutions to complex reasoning problems. An effective technique to enhance LLM performance is  in-context learning, which encourages a step-by-step reasoning process by including explanatory examples to guide the model's responses. However, selecting appropriate exemplars for the model poses a challenge, as each dataset demands a distinct set of exemplars to enable the LLM to learn effectively and perform well on the test set. Current studies often rely on uncertainty- or diversity-based selection strategies to select exemplars for annotation and to improve model learning. However, these studies typically employ a non-adaptive approach, selecting a set of exemplars all at once. We argue that this non-adaptive strategy may result in a set of exemplars with high redundancy in terms of the knowledge covered, ultimately reducing their overall informativeness. To address this limitation, we propose \textsc{Adaptive-Prompt}, a novel method that adaptively selects exemplars by leveraging model feedback from previously chosen exemplars. Experimental results show that \textsc{Adaptive-Prompt} significantly enhances LLM performance across a variety of reasoning tasks.
\end{abstract}
\section{Introduction}

Large language models (LLMs) have demonstrated their exceptional proficiency across a broad spectrum of tasks and fundamentally transformed the field of natural language processing (NLP) \cite{kojima2022large,wei2022emergent,shao2023synthetic,achiam2023gpt,touvron2023llama,zhao2023survey}. However, LLMs often struggle with tasks that require complex reasoning. In-context learning (ICL) is a powerful solution to this issue by guiding models through examples and instructions without modifying their parameters. This approach is both convenient and effective, making it increasingly popular in various applications.

One approach within the ICL framework is Chain-of-Thought (CoT) prompting \cite{wei2022chain}, which has proven highly effective by breaking down complex problems into sequential, step-by-step explanations. A variant known as zero-shot CoT \cite{kojima2022large} involves appending the phrase ``Let's think step by step'' to the end of a question, allowing the model to generate reasoning without any demonstrations. In contrast, few-shot CoT \cite{wei2022chain} provides a small set of question-rationale-answer demonstrations before the test question, thereby guiding the model in how to approach and solve the problem. However, the traditional few-shot CoT approach is constrained by its dependence on a fixed set of human-crafted exemplars, which may not always be optimal for the test set.

Recent advancements in ICL have introduced several approaches, including Auto-CoT \cite{zhang2022automatic}, Active-Prompt \cite{diao2023active}, and  ADAICL \cite{mavromatis2023examples}, among others.  Auto-CoT utilizes vector representations of questions to cluster them and then sample questions from each cluster based on their proximity to the cluster center, ensuring diversity within the exemplar set. Active-Prompt, on the other hand, ranks questions in the training set by their uncertainty in a one-shot manner, and a subset of the most uncertain questions is selected for annotation. Similarly,  ADAICL leverages LLM feedback to identify the most uncertain questions, divides them into distinct regions, and selects the most representative question from each region. These methods aim to address the limitations of fixed exemplar sets by incorporating either uncertainty- or diversity-based selection strategies to enhance model performance across various reasoning tasks. Our work builds upon and extends Active-Prompt \cite{diao2023active}. While Active-Prompt has made significant progress and demonstrated effectiveness, it may introduce redundancy by overlooking the similarity between the selected exemplars. Since models often vary in their ability to handle different types of questions \cite{zhang2022automatic}, focusing solely on the most uncertain questions can lead to clusters around similar problem types, potentially under-representing the diversity of the task space. To address this, we introduce a novel approach called Adaptive Chain-of-Thought Prompting (\textsc{Adaptive-Prompt}). Consistent with existing studies in ICL, our goal is to identify a small set of the most informative questions for annotation.  \textsc{Adaptive-Prompt} operates adaptively and iteratively: in each iteration, it selects the most uncertain question, given the previously selected and annotated exemplars, to add to the exemplar set. Thus, each exemplar's selection depends on all previously chosen exemplars, in contrast to Active-Prompt, where exemplars are selected independently. By incorporating principles from adaptive learning and subset selection, our approach maintains exemplar diversity while enhancing LLM performance.

We evaluated the effectiveness of our \textsc{Adaptive-Prompt} method through extensive experiments on arithmetic, commonsense, and symbolic reasoning tasks. The results demonstrate that our approach outperforms existing uncertainty- and diversity-based strategies.

\section{Related Work}

\paragraph{Chain-of-thought Prompting}
LLMs demonstrate enhanced capabilities with CoT methods \cite{wei2022chain}. The fundamental idea is to have the model explicitly generate intermediate steps in reasoning before arriving at the final answer, formatted as $\langle$ Question, Reasoning Chain, Answer $\rangle$. Zero-shot learning and few-shot learning have been foundational in traditional language models. The zero-shot method does not provide any examples or guidance but simply prompts the model with the phrase ``Let's think step by step.'' In contrast, the few-shot CoT method provides the model with a set of exemplars, including the question, rationale, and answer. Since the introduction of CoT, a series of studies and experiments have explored the effectiveness of these methods in various contexts. Self-Consistency \cite{wang2022self} builds on CoT by generating multiple reasoning paths for the same question, with the final answer determined through a voting mechanism among these paths. Auto-CoT \cite{zhang2022automatic} uses the distance of questions' embeddings in latent space to further automate the CoT process by dynamically selecting and constructing exemplars. In few-shot CoT, identifying an optimal exemplar set is crucial since the effectiveness of ICL highly depends on it. Employing an intuitive similarity-based selection strategy, \cite{liu2021makes} use the nearest neighbors of a given test sample in the embedding space as exemplars. This method however, is computationally expensive, as it must be performed for each test sample. Another strategy is to construct a fixed set of examplars that can be applied to the whole test set. The criterion can be the complexity of reason chain \cite{fu2022complexity}, fairness of the model's predictive distribution \cite{ma2023fairness} and/or uncertainty of the model's generated answers \cite{diao2023active}. Zhou et al. \cite{zhou2023inform} use the information entropy of outcomes to rank the questions and apply a diversity filter to enhance the diversity of the exemplar set. Mavromatis et al. \cite{mavromatis2023examples} consider exemplar selection as a maximum coverage problem. They first identify the most uncertain examples using the model's feedback, then define a region for each uncertain example and select the most representative question within each region as an exemplar.

\paragraph{Active Learning in NLP}
Active learning strategies have long been recognized in the NLP community as an effective approach to reducing annotation costs while maintaining or even improving model performance \cite{zhu2008active, settles2009active}. It aims to select the most informative examples, often using criteria such as model uncertainty, expected information gain, or diversity to maximize the efficiency. Researchers have explored active learning in various NLP tasks, such as text classification \cite{schroder2020survey}, text recognition \cite{erdmann2019practical}, and machine translation \cite{zhao2020active}. In the context of LLM, there is a series of studies that use active learning to fine-tune LLM \cite{margatina2021active,schroder2021revisiting,koksal2022meal}. Another direction for applying active learning is through LLMs' in-context learning \cite{su2022selective, diao2023active, mavromatis2023examples}.

\section{Problem Formulation}
The input of our problem consists of a LLM $M$, a training set of $m$ unlabeled questions $Q$, i.e., $Q = \{ q_1, q_2, \dots, q_m \}$, and a test set $P$ containing $n$ questions, i.e., $P = \{ p_1, p_2, \dots, p_n \}$. Our objective is to select $k$ questions from $Q$, denoted as $\{ q_1, q_2, \dots, q_k \}$, and annotate them to create an \emph{exemplar set} $E = \{ (q_1, r_1, a_1), (q_2, r_2, a_2), \dots, (q_k, r_k, a_k) \}$, where for each $i \in \{ 1, 2, \dots, k \}$, $r_i$ represents the reasoning chain and $a_i$ is the correct answer for question $q_i$. Here $k$ is a budget constraint, limiting the maximum size of the exemplar set.

The goal is to identify an optimal $E$ such that, when any test question $p$ from $P$ is presented to $M$ along with $E$, the model generates the most accurate  response for $p$.
\begin{figure*}[t]
    \centering
    \includegraphics[width=\linewidth]{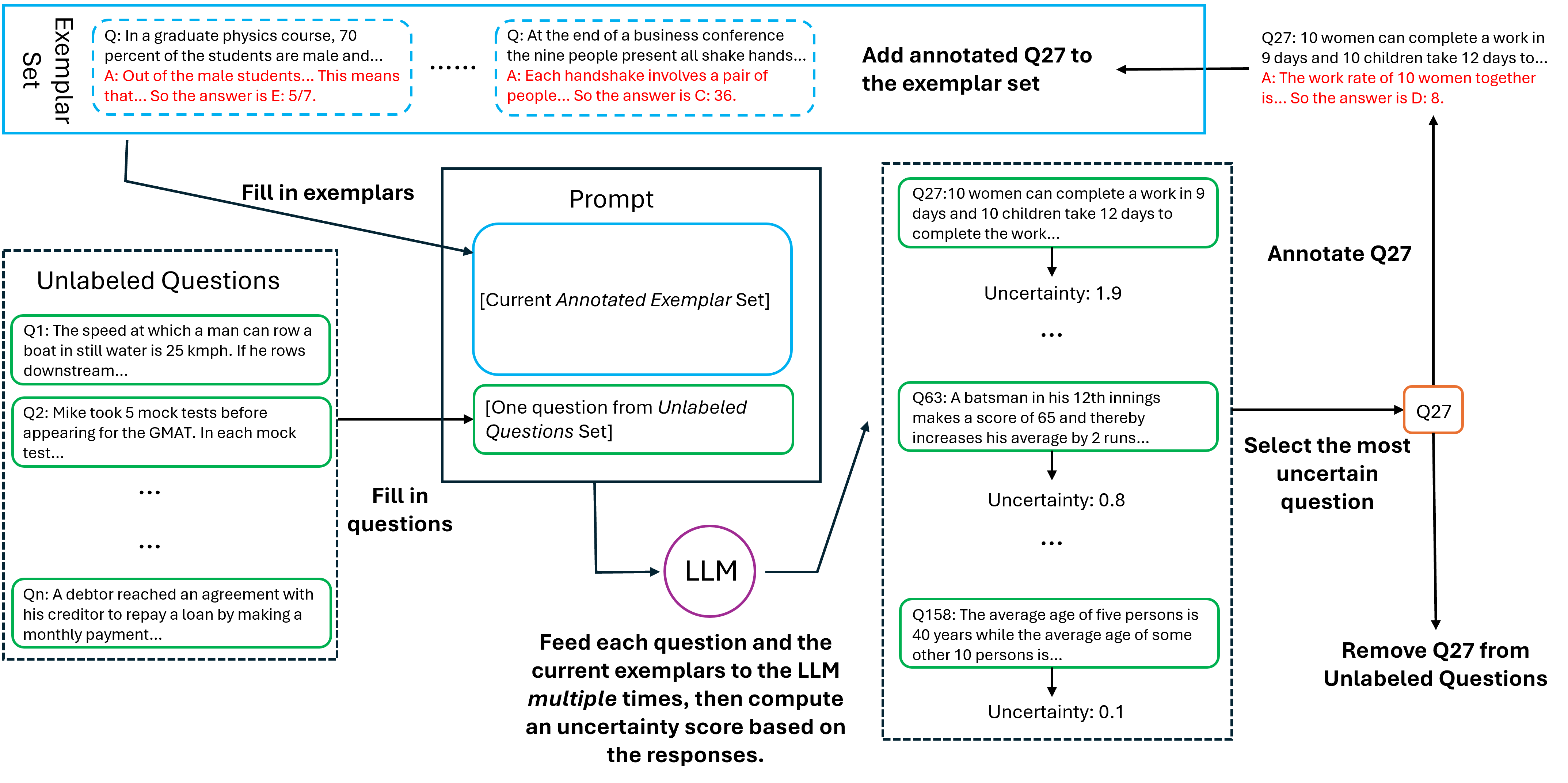}
    \caption{Illustration of \textsc{Adaptive-Prompt}.}
    \label{fig:Diagram}
\end{figure*}
\section{Design of Adaptive-Prompt}
In this paper, we introduce a novel in-context learning framework, \textsc{Adaptive-Prompt}. An illustration of \textsc{Adaptive-Prompt} is presented in Figure \ref{fig:Diagram}, with a detailed implementation provided in Algorithm \ref{alg:adaptive_prompt}.

\begin{algorithm}
\caption{\textsc{Adaptive-Prompt}}\label{alg:adaptive_prompt}
\begin{algorithmic}[1]
    \STATE \textbf{Input:} Unlabeled questions $Q$, size constraint $k$, a large language model $M$
     \STATE \textbf{Output:} Exemplar set $E$ of size $k$
    \STATE Initialize exemplar set $E = \emptyset$
    \WHILE{$|E| \leq k$}
        \FOR{each question $q_i \in Q$}
            \STATE Feed $E \oplus q_i$ to $M$  $l$ times and compute the uncertainty score $u(q_i\mid E)$
        \ENDFOR
        \STATE Select the question $q_j \in Q$ with the highest uncertainty, i.e., $q_j \in \arg\max_{q\in Q} u(q \mid E)$
        \STATE Manually annotate $q_j$ with reasoning chain $r_j$ and answer $a_j$
        \STATE  $E \leftarrow E \cup (q_j, r_j, a_j)$, $Q\leftarrow Q\setminus \{q_j\}$
    \ENDWHILE
    \RETURN $E$
\end{algorithmic}
\end{algorithm}
\subsection{Design of \textsc{Adaptive-Prompt}}
We next give a detailed description of \textsc{Adaptive-Prompt}. The basic idea of this approach is to \emph{adaptively} select a group of \emph{representative} exemplars  to help improve the reasoning capabilities of LLMs. The process consists of selecting the most uncertain questions, annotating them, and recalculating uncertainty scores, so that the model can continuously leverage the updated exemplar set to do uncertainty estimation. The steps are as follows:

\textbf{1. Initialization}: Set the exemplar set $E$ as an empty set, $E = \emptyset$.

\textbf{2. Uncertainty Evaluation}: For each question $q_i \in Q$, a prompt is constructed by combining the current exemplar set $E$ with the question $q_i$. Let $E\oplus q_i$ denote such a prompt. We perform $l$ independent queries by feeding $E \oplus q_i$ to the LLM and collect $l$ corresponding responses. To capture the variance among those $l$ responses, we compute an uncertainty score $u(q_i\mid E)$. A detailed definition of $u(q_i\mid E)$ can be found in Section \ref{sec:uncertainty}. Intuitively, we use $u(q_i\mid E)$ to measure the extent of the LLM's confidence in providing a correct answer to $q_i$, given that $E$ has been provided.

\textbf{3. Selecting the Most Uncertain Question for Annotation}: Once uncertainty scores $u(q_i\mid E)$ are computed for all questions in $Q$, the questions are ranked by their uncertainty scores. The question with the highest uncertainty, $q_j$, is selected for annotation. The selected question $q_j$ is manually annotated with a reasoning chain $r_j$ and an answer $a_j$. The annotated exemplar $(q_j, r_j, a_j)$ is added to the exemplar set $E$ and the question $q_j$ is removed from the training set $Q$. That is, we update the exemplar set $E$ and the unlabeled question set $Q$ as follows: $E \leftarrow E \cup (q_j, r_j, a_j)$, $Q\leftarrow Q\setminus \{q_j\}$. The intuition behind this selection rule is that if a question $q_j$ has the highest uncertainty score, it indicates that the current exemplar set $E$ does not provide sufficient knowledge for the LLM to confidently answer $q_j$. In this case, adding $q_j$ to $E$ is expected to best expand the LLM's knowledge base.

       During the annotation process, annotators must carefully read each question and break down the reasoning into clear, incremental steps that lead to the final answer. To ensure consistency, all annotations within a single experiment are provided by the same human annotator.

\textbf{4. Iterative Evaluation and Selection}: Steps 2 -- 3 are repeated iteratively. After each update to $E$, the uncertainty scores for the remaining questions in $Q$ are recalculated based on the updated exemplar set $E$. That is, this process starts as zero-shot prompting when $E = \emptyset$, and in subsequent iterations, the model performs 1-shot, 2-shot, and so on, incorporating the expanding exemplar set into the prompts. This iterative process continues until the exemplar set reaches the desired size $k$.

\paragraph{Comparison with Active-Prompt\cite{diao2023active}} Our approach distinguishes itself from other ICL approaches by adopting an \emph{adaptive} selection rule to expand the exemplar set. In our framework, the selection of the exemplar set is performed iteratively and adaptively, with the next exemplar chosen based on all previously selected exemplars. In contrast, most existing studies in this field use a non-adaptive selection rule. For example, Active-Prompt \cite{diao2023active}, the framework our study builds upon, employs a non-adaptive selection rule. Their method computes the uncertainty score for each question all at once, assuming a zero-shot learning scenario. Specifically, Active-Prompt first calculates $u(q_i \mid E = \emptyset)$ for all questions $q_i \in Q$ and then selects the top $k$ questions with the highest uncertainty to annotate. Here, $\emptyset$ refers to either an empty set or a few human-written chain-of-thoughts. One potential limitation of this non-adaptive approach is that it may fail to  capture redundancy among selected exemplars. For instance, consider two questions $q_1 = \{\text{What is the current world population?}\}$ and $q_2 = \{\text{What is the number of people living on Earth?}\}$. If we compute the uncertainty scores for both questions independently, $u(q_1 \mid E = \emptyset)$ and $u(q_2 \mid E = \emptyset)$, it is possible that both questions yield very high uncertainty scores, causing Active-Prompt to select both for annotation. However, it is easy to verify that both questions essentially cover the same knowledge point, making it redundant to annotate both. Now, consider our  \textsc{Adaptive-Prompt}: if $q_1$ has been selected previously, then $u(q_2 \mid E = \{q_1\})$ will likely have a low value. Thus, our adaptive approach successfully selects only one of the two questions for annotation, avoiding redundancy and enhancing the informativeness of the exemplar set.

\subsection{Computing the Uncertainty Score $u(q\mid E)$}
\label{sec:uncertainty}
Recall that a critical step in \textsc{Adaptive-Prompt} is calculating the uncertainty score $u(q \mid E)$ for a given exemplar set $E$ and question $q$. This score, $u(q \mid E)$, quantifies the potential divergence in the model's responses \cite{gentile2024fast}. Intuitively, a higher $u(q \mid E)$ indicates lower model confidence in answering $q$ based on the exemplar set $E$. Following \cite{diao2023active}, we use two metrics to define this score: \emph{disagreement} and \emph{entropy}. Alternative forms of uncertainty scores such as variance could also be developed to capture additional aspects of model confidence.

For a given exemplar set $E$ and a question $q$, consider performing $l$ independent queries by presenting the combined input $E \oplus q$ to the LLM, resulting in a set of $l$ responses, denoted by $A = \{a_1, a_2, \dots, a_l\}$. Here, each response $a_i$ is generated independently based on the exemplar-question pair $E \oplus q$. To identify distinct answers, let $A_u = \{a_1', a_2', \dots, a_t'\}$ represent the set of unique responses in $A$, where $t \leq l$ and each $a_i'$ is a distinct response found among the generated answers. This set $A_u$ captures the diversity of responses provided by the model when given $E$ and $q$. The disagreement-based uncertainty score is defined as $u(q \mid E) = \frac{t}{l}$, representing the ratio of distinct answers $t$ to the total number of generated answers $l$. This score reflects the proportion of unique responses produced by the model, indicating its variability in answering $q$ given the exemplar set $E$. The entropy-based uncertainty score is defined as $u(q \mid E) = -\sum_{j=1}^{t} P(a_j') \log P(a_j')$, where $P(a_j')$ represents the frequency of a unique answer $a_j'$ in the set of unique responses $A_u$.  A higher entropy value indicates a greater uncertainty in the model’s predictions.

\section{Experimental Settings}
We conduct a series of experiments on several public datasets using the backbone generative models GPT-3.5 Turbo and GPT-4o Mini \cite{openai-models}, which are renowned for their robust natural language understanding and advanced reasoning capabilities. Since both GPT-3.5 Turbo and GPT-4o Mini are closed-source, we access them through the OpenAI API.  The details of the datasets, models, and experimental setups are provided below.

\paragraph{Datasets and Models}
We adopt the evaluation settings from previous reasoning studies and conduct experiments on three task categories: arithmetic reasoning, commonsense reasoning, and symbolic reasoning.
For arithmetic reasoning, we use 3 datasets: GSM8K \cite{cobbe2021training}, SVAMP \cite{patel2021nlp} and AQuA \cite{ling2017program}.
For commonsense reasoning, we use 2 datasets: StrategyQA \cite{geva2021did} and CSQA \cite{talmor2019commonsenseqa}. And we use one symbolic reasoning task dataset: Last letter concatenation (Letter Concat) \cite{wei2022chain}. 

\paragraph{Baselines}
We compare the performance of our method with several baselines: Zero-Shot CoT \cite{kojima2022large}, (Few-Shot) CoT \cite{wei2022chain}, Auto-CoT \cite{zhang2022automatic}, Random-CoT and Active-Prompt \cite{diao2023active}. Random-CoT selects questions from the training set at random and then annotates them. Among the baselines, Auto-CoT stands out as a diversity-based approach. It clusters questions from the dataset based on the similarity of their embeddings, then selects a representative question from each cluster. For these representative questions, it generates reasoning chains using the Zero-Shot CoT method. On the other hand, Active-Prompt employs a non-adaptive, uncertainty-based strategy that identifies a set of the most uncertain questions all at once.  In contrast, our approach adopts an adaptive selection rule to expand the exemplar set.

\paragraph{Settings}
We follow the settings in \cite{wei2022chain} and set the number of exemplars $k$ for each dataset as follows: 4 for AQuA and Letter Concat, 6 for StrategyQA, 7 for CSQA, and 8 for GSM8K and SVAMP. The SVAMP dataset does not have a training set, so we use the exemplar set from GSM8K for inference, as both datasets involve mathematical reasoning tasks with similar formats.  For Letter Concat, we utilize an out-of-distribution test set with four-letter concatenations, while the training set contains three-letter concatenations. The number of times a question is answered, $l$, is set to 10 to be consistent with the previous studies \cite{diao2023active}. We also conduct experiments to investigate the impact of different values of $k$ on the performance. For datasets with very large training sets, it is computationally expensive for the model to answer all the questions in the training set. To address this, we select a subset of questions for adaptive selection. In our experiments, we set the maximum size of the candidate question pool to $s = 50 \times$[number of exemplars $k$] (e.g., 200 for AQuA, 300 for StrategyQA, and 400 for GSM8K). If the training set size exceeds $s$, we randomly select $s$ questions to form the training subset used in the experiment. If the training set is smaller than or equal to $s$, we use the entire training dataset. The performance of the LLM on the test set is subject to significant variability, which arises from several sources of randomness: the selection of the training question set, the model's responses during exemplar set construction, and the model's answers to the test questions. To mitigate this randomness and improve the robustness of the evaluation, we repeat the entire selection process-including the selection of training questions and exemplar set construction-three times for each dataset. For each constructed exemplar set, we adopt the Self-Consistency approach \cite{wang2022self}: the model answers each test question six times, and the most frequent answer is used to determine the accuracy. The final accuracy is then computed as the average of the accuracy values obtained from the three experimental runs.

\begin{table*}[t]
    \centering
    \caption{Performance Comparison on GPT-3.5 Turbo}
    \label{Table: GPT3.5}
    \begin{tabular}{lcccccccc}
        \toprule
        & \multicolumn{3}{c}{Arithmetic} & \multicolumn{2}{c}{Commonsense} & \multicolumn{1}{c}{Symbolic}\\
        \cmidrule(lr){2-4} \cmidrule(lr){5-6} \cmidrule(lr){7-7}  \\
        & AQuA & GSM8K & SVAMP & StrategyQA & CSQA & Letter Concat & Average\\
        \midrule
        Zero-Shot CoT & \textbf{62.7} & 80.4 & 81.3 & 70.5 & 76.2 & 52.6 & 70.6\\
        CoT & 59.3 & 81.2 & 80.7 & 74.1 & 74.8 & 73.5 & 73.9\\
        Auto-CoT & 58.9 & 81.6 & 81.7 & 74.9 & 76.0 & 74.0 & 74.5\\
        Random-CoT & 60.6 & 81.5 & 80.6 & 74.2 & 75.6 & 73.7 & 74.3\\
        Active-Prompt (D) & 60.6 & 81.9 & 81.8 & 76.0 & 76.9 & 74.5 & 75.3\\
        Active-Prompt (E) & 60.8 & 81.8 & 82.4 & 76.0 & 77.2 & 73.8 & 75.3\\
        Adaptive-Prompt (D) & 61.9 & 82.0 & \textbf{82.5} & \textbf{76.6} & 77.5 & 74.4 & 75.8\\
        Adaptive-Prompt (E) & 62.2 & \textbf{82.7} & 82.2 & 76.4 & \textbf{77.7} & \textbf{74.7} & \textbf{76.0}\\
        \bottomrule
    \end{tabular}
\end{table*}
\begin{table*}[t]
    \centering
    \caption{Performance Comparison on GPT-4o mini}
    \label{Table: GPT4o}
    \begin{tabular}{lcccccccc}
        \toprule
        & \multicolumn{3}{c}{Arithmetic} & \multicolumn{2}{c}{Commonsense} & \multicolumn{1}{c}{Symbolic}\\
        \cmidrule(lr){2-4} \cmidrule(lr){5-6} \cmidrule(lr){7-7}  \\
        & AQuA & GSM8K & SVAMP & StrategyQA & CSQA & Letter Concat & Average\\
        \midrule
        Zero-Shot CoT & 81.5 & 93.7 & \textbf{94.1} & 76.9 & 80.5 & \textbf{91.1}* & 86.3* \\
        CoT & \textbf{82.3} & 93.1 & 93.2 & 77.9 & 80.8 & 87.7 & 85.8\\
        Auto-CoT & 82.0 & 93.8 & 93.4 & 78.4 & 81.2 & 86.7 & 85.9\\
        Random-CoT & 81.6 & 93.6 & 92.8 & 78.3 & 81.5 & 88.0 & 86.0\\
        Active-Prompt (D) & 81.7 & 93.6 & 93.2 & 79.0 & 82.8 & 88.2 & 86.4\\
        Active-Prompt (E) & 82.1 & 93.7 & 93.2 & 79.7 & 83.3 & 88.1 & 86.7\\
        Adaptive-Prompt (D) & 82.0 & \textbf{94.2} & 93.7 & \textbf{80.3} & 83.1 & 88.1 & \textbf{86.9}\\
        Adaptive-Prompt (E) & \textbf{82.3} & \textbf{94.2} & 93.3 & 79.7 & \textbf{83.6} & 88.3 & \textbf{86.9}\\
        \bottomrule
    \end{tabular}
\end{table*}
\section{Experiment Results}

The performance of our method and the baseline approaches on GPT models is presented in Table \ref{Table: GPT3.5} and \ref{Table: GPT4o}. All results are expressed as percentages, with the best outcomes highlighted in \textbf{bold}.

When using GPT-3.5 Turbo, our method achieves the best performance on 5 out of 6 datasets. The one exception is the AQuA dataset, where the Zero-Shot CoT method achieves an accuracy of $62.7\%$ and outperforms all other approaches including ours. On two other arithmetic datasets,  \textsc{Adaptive-Prompt} attains the highest accuracy compared to all baseline models. On GSM8K, the marginal increase over the most competitive baseline, Active-Prompt (D), is 0.8\% with the entropy-based variant. On SVAMP, the disagreement-based variant achieves a gain of 0.1\%. For the commonsense datasets, \textsc{Adaptive-Prompt} increases the accuracy by $0.6\%$ on StrategyQA using entropy as the metric, while the gain on CSQA is $0.5\%$ compared to best-performing baseline model. In the Letter Concatenation task, all few-shot CoT methods outperform the Zero-Shot CoT by a significant margin (over a $20\%$ of increase), and our approach improves the performance by an additional $0.2\%$. Regarding the average accuracy across all evaluated tasks, the entropy-based \textsc{Adaptive-Prompt} improves by $0.7\%$ compared to Active-Prompt using the same metric.

The outcomes obtained with GPT-4o mini show a distinct difference. For arithmetic tasks, on the AQuA dataset, the highest accuracy is achieved by both CoT and the entropy-based \textsc{Adaptive-Prompt}. On GSM8K, our method improves performance by $0.5\%$ compared to the best baseline model. However, on SVAMP, Zero-Shot CoT outperforms all few-shot learning approaches. On two commonsense tasks, StrategyQA and CSQA, \textsc{Adaptive-Prompt} increases accuracy by $0.6\%$ and $0.3\%$ respectively. Additionally, on the Letter Concatenation task, the Zero-Shot CoT now achieves the highest accuracy, exceeding $90\%$, which contrast the much poorer results from GPT-3.5 Turbo. Our method achieves slightly higher accuracy than other few-shot CoT approaches, thereby narrowing the gap with Zero-Shot CoT. One possible explanation for this result is that the GPT-4o mini has gained the ability to split words into individual letters rather than treating them as single tokens (whole words). We find that \textsc{Adaptive-Prompt} outperforms the baseline models on the majority of datasets across both GPT-3.5 Turbo and GPT-4o mini, demonstrating its effectiveness. Moreover, we observe that the performance gains achieved by our method are more substantial on GPT-3.5 Turbo compared to GPT-4o mini. This suggests that our approach serves as a complementary enhancement to LLMs; as the inherent performance of LLMs improves, the incremental benefits provided by \textsc{Adaptive-Prompt} tend to diminish.

\paragraph{Effect of Annotators}
It is intuitive that the style and quality of annotations will significantly impact the outcomes of experiments. In this section, we examine the effects of different human annotators. To do so, we have a second annotator provide annotations for the selected questions while keeping all other processes the same. Here, we only adopt the entropy-based approach. We conduct experiments on the following datasets: GSM8K, StrategyQA, and CSQA, using GPT-3.5 Turbo.
The results are presented in Table \ref{Table:Annatator}. The first three rows do not require human annotation, while the last three are based on annotations from a different annotator. The results align with those from the first annotator: \textsc{Adaptive-Prompt} either matches or exceeds the performance of all other baselines.

\begin{table}[t]
    \caption{}
    \centering
            \label{Table:Annatator}
    \begin{tabular}{l|c|c|c}
        & GSM8K & StrategyQA & CSQA \\
        \hline
        Zero-Shot CoT & 80.4 & 70.5 & 76.2\\
        CoT & 81.2 & 74.1 & 74.8\\
        Auto-CoT & 81.6 & 74.9 & 76.0\\
        \hline
        Random CoT & 81.9 & 73.9 & 76.2\\
        Active-Prompt (E) & 82.1 & \textbf{76.7} & 77.0\\
        Adaptive-Prompt (E) & \textbf{82.5} & \textbf{76.7} & \textbf{77.3}\\
    \end{tabular}
\end{table}
\begin{figure}[t]
\hspace{-1.1cm}
    \centering
    \begin{minipage}{0.45\textwidth}
        \centering
        \includegraphics[scale=0.3]{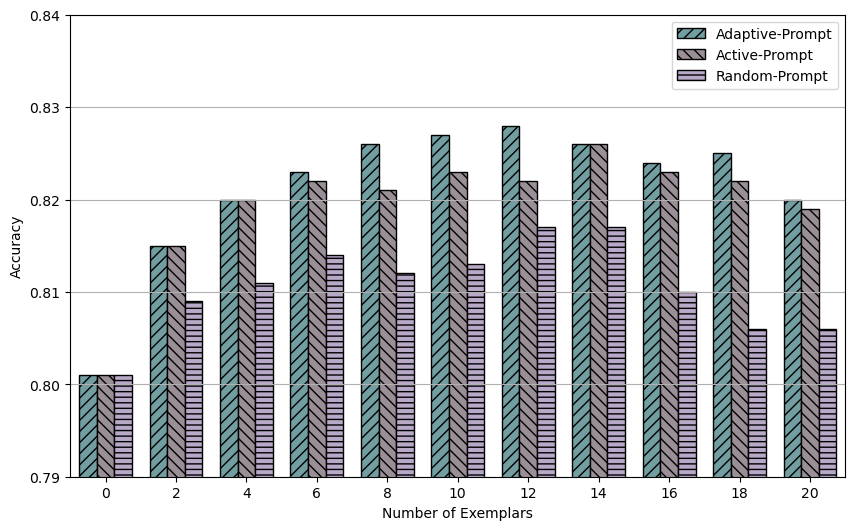}
        \caption{GSM8K}
        \label{fig:GSM8K}
    \end{minipage}
    \hfill
    \begin{minipage}{0.45\textwidth} \hspace{-1.1cm}
        \centering
        \includegraphics[scale=0.3]{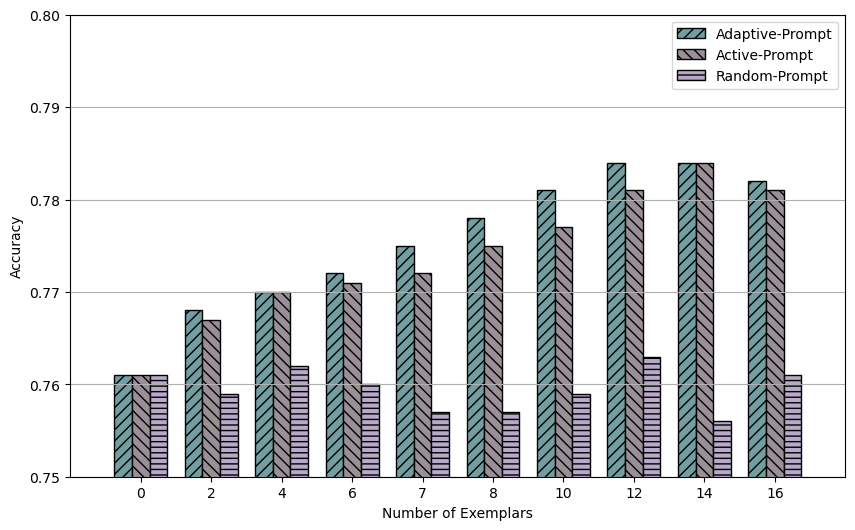}
        \caption{CSQA}
        \label{fig:CSQA}
    \end{minipage}
\end{figure}
\paragraph{Effect of Exemplar Set Size}
The size of the exemplar set, $k$, is a critical factor influencing the performance of LLMs within the framework of ICL. A small exemplar set may fail to provide sufficient and representative information for the model to fully utilize its ICL potential. Conversely, an excessively large exemplar set risks surpassing the model's capacity to handle long input sequences effectively. Also, the effectiveness of our method is influenced by the value of $k$. When $k$ is too small, there are insufficient exemplars to fully exploit the benefits of adaptive selection, resulting in performance comparable to a one-shot strategy. Conversely, when $k$ is too large, all baseline methods include enough exemplars to cover nearly all case categories, diminishing the relative advantage of our approach. Thus, identifying an optimal range for $k$ is crucial to maximizing the effectiveness of our method and the LLM's performance. In this section, we examine the impact of  $k$ on model performance, using the GSM8K and CSQA datasets. The maximum number of exemplars $k$ is increased to 16 for GSM8K and 20 for CSQA, with results reported at intervals of 2. Additionally, for CSQA, we include the results for $k=7$ from the main experiments. The outcomes of an experiment run can be found in Figure \ref{fig:GSM8K} and \ref{fig:CSQA}. We compare the performance of our method with Active-Prompt,  the most relevant approach to ours. Additionally, we include Random-Prompt as a baseline for comparison, as shown in the charts. For GSM8K, as the number of exemplars increases, the accuracy of all three methods initially rises but subsequently declines slightly. But for CSQA, Random-Prompt does not benefit from the growing exemplar set. Our method consistently outperforms both baselines across tasks, although at certain points, Active-Prompt reaches the same accuracy as our method. The largest gap between Active-Prompt and our method on the measured data occurs when $k=12$ for GSM8K and $k=10$ for CSQA.
This result further confirms our hypothesis and the conclusions drawn from the main experiments.

\paragraph{Evaluation with Weaker Models}
We conduct additional evaluations using the LLaMA3-8B model, known for its cost-effectiveness. In most cases, Zero-Shot CoT outperforms both baselines and our method. This may be due to weaker models struggling with long chains of thought, as longer prompts can degrade performance \cite{liu2024lost}.

\bibliographystyle{splncs04}
\bibliography{main}
\end{document}